# Research on Short-Video Platform User Decision-Making via Multimodal Temporal Modeling and Reinforcement Learning

Jinmeiyang Wang[1*], Jing Dong[2], Li Zhou[3]
[1]School of Media & Communication, Shanghai Jiaotong University, Shanghai, 200240, China.
[2]Columbia University, New York, 10027, US.
[3]McGill University, Montréal, 27708, Canada.
*Corresponding author Email: wangjmy2000@163.com

## ABSTRACT

This paper proposes the MT-DQN model, which integrates a Transformer, Temporal Graph Neural Network (TGNN), and Deep Q-Network (DQN) to address the challenges of predicting user behavior and optimizing recommendation strategies in short-video environments. Experiments demonstrated that MT-DQN consistently outperforms traditional concatenated models, such as Concat-Modal, achieving an average F1-score improvement of 10.97% and an average NDCG@5 improvement of 8.3%. Compared to the classic reinforcement learning model Vanilla-DQN, MT-DQN reduces MSE by 34.8% and MAE by 26.5%. Nonetheless, we also recognize challenges in deploying MT-DQN in real-world scenarios, such as its computational cost and latency sensitivity during online inference, which will be addressed through future architectural optimization.

## 1 Introduction

With the rapid development of mobile Internet and digital technology, short video platforms have become one of the social media applications with the largest user scale and the highest level of activity in the world. Platforms represented by Douyin, TikTok, and Kuaishou have more than 600 million daily active users, and the average daily usage time of users generally exceeds 90 minutes(He et al., 2024; Jiang et al., 2022). Their powerful content dissemination and social influence are profoundly reshaping the way of information consumption and the digital cultural ecology. However, as the platform content becomes increasingly saturated and user interests tend to be diverse and changeable, the platform faces severe challenges in user retention and content distribution efficiency(Jing & Qing, 2024; Ping & Yue, 2024). Relevant data show that the 7-day retention rate of new users is generally lower than 25%, while the recommendation click rate of active users continues to decline. Some users have high screen-sliding bounce rates and weakened willingness to interact(Jiang et al., 2022; Sannidhan et al., 2023). These problems reflect that the current recommendation system has obvious bottlenecks in understanding and predicting user behavior. In this context, in-depth understanding of users' decision-making behavior on short video platforms (such as content selection, interactive participation, social sharing, etc.) is not only a key fulcrum for improving recommendation effects, improving user experience and extending the life cycle of the platform, but also an important academic topic to reveal the information dissemination mechanism of social media and release the potential of the digital economy.

While existing research has predominantly applied traditional machine learning techniques or unimodal data analyses to investigate user behavior—such as employing text mining for sentiment analysis of comments or predicting video popularity based solely on visual features—these approaches exhibit notable limitations(Cai et al., 2022). Short videos inherently combine visual, audio, and textual elements, making unimodal analysis insufficient to fully grasp the content's semantic and emotional depth(Goodwin et al., 2024; Zhou et al., 2021). The user decision-making process is inherently dynamic and sequential, and is influenced by historical behavior, social relationships, and real-time context. Traditional models cannot effectively model the dynamic evolution of temporal dependencies and user preferences.

Multimodal deep learning can more comprehensively characterize the content characteristics of short videos by integrating heterogeneous data such as images, texts, and audios(He & Li, 2024; Xing et al., 2025). Reinforcement learning, focused on dynamic decision-making, optimizes strategies through interactions between an agent and its environment, making it ideal for simulating user decisions on platforms(He et al., 2025; Mubarak et al., 2021). The challenge remains to effectively combine multimodal data processing, temporal modeling, and reinforcement learning to develop intelligent models that accurately reflect user decision-making(He et al., 2021; Rezaee et al., 2024).

This paper introduces the MT-DQN (Multimodal Temporal Deep Q-Network) model as a solution to these challenges, focusing on the analysis of user decision-making behaviors on short-video platforms. The proposed model is comprised of three key components: a Transformer-based multimodal fusion module (Zhao et al., 2023) that enables cross-modal semantic understanding of video content; a temporal graph neural network module (Su & Wu, 2025) designed to capture the evolving patterns of user behavior and social interactions; and a deep Q-network module (Yan et al., 2021) that refines decision-making strategies through reinforcement learning. Together, these elements work in tandem to create a robust framework for user behavior analysis, addressing content perception, behavior modeling, and decision optimization. The contributions of this work are summarized as follows:

1) Innovative Model Architecture: This study overcomes the limitations of traditional user behavior models, which typically handle multimodal data and temporal dynamics separately.

2) Theoretical and Methodological Advancements: In the field of multimodal learning, we propose adaptive temporal attention mechanisms and cross-modal feature fusion strategies that effectively address the intrinsic heterogeneity of multimodal data and the intricate challenges posed by temporal dependency modeling. Within reinforcement learning applications, the design of a dynamic reward function, tailored to reflect users' long-term value, significantly enriches the theoretical underpinnings of reinforcement learning-based decision frameworks in social media contexts, thereby broadening the methodological landscape for studying user decision-making behavior.

3) Practical Application Value: The proposed MT-DQN model exhibits high transferability and strong practical potential for large-scale deployment in real-world scenarios. Its research outcomes provide essential technical support for optimizing personalized recommendation systems, informing strategic content creation, and formulating user growth initiatives on short-video platforms. These contributions empower platforms to enhance operational efficiency, boost user engagement, and improve overall user experience.

The remainder of this paper is structured as follows. Section 2 reviews related research on multimodal deep learning and reinforcement learning in the context of user behavior analysis. Section 3 details the model construction methodology and technical implementation. Section 4 presents experimental validation of the model's performance, followed by a discussion of the results and potential directions for future research.

## 2 Related Work

**2.1 Multimodal Behavior Analysis**

The development of multimodal fusion technology has experienced a gradual evolution from early shallow splicing to deep attention mechanism, providing strong technical support for social media user behavior analysis. In the early stage, researchers mostly used feature concatenation or weighted average fusion methods to simply concatenate or average the features of modalities such as images, texts, and audios and input them into the prediction model to predict basic behaviors such as user likes and comments(Jabarin et al., 2022; Zhang et al., 2022). This type of method has high computational efficiency, but weak ability to mine semantic relationships between modalities, which easily causes information redundancy and semantic conflicts, resulting in poor performance of the model when facing complex behaviors(Pelachaud et al., 2021; Sümer et al., 2021). After entering the deep learning stage, the fusion method gradually evolved to alignment-based fusion and gated mechanism, such as using bidirectional RNN to collaboratively model image descriptions and user behavior sequences, or using gated neural networks to control the flow and retention of information in each modality(Umer & Sharif, 2022). Although the fusion strategy at this stage has improved the expressiveness of the model to a certain extent, the modeling of modal heterogeneity and temporal dynamics is still insufficient. The introduction of the Transformer architecture marks that multimodal fusion has entered the stage of deep fusion driven by self-attention(Zheng et al., 2024). Model frameworks represented by BERT and ViT use the self-attention mechanism to achieve interactive alignment between modalities, adaptive weight learning, and context-enhanced representation(Li et al., 2025). For example, the Multimodal Attention Network (MAN) accurately captures the local area of interest to users by calculating the attention distribution between visual frames and text keywords, effectively improving the discrimination accuracy of recommendation systems and user behavior analysis models. On this basis, the introduction of the Cross-Modal Attention mechanism further promotes the deep coupling between images, texts, and sounds, and builds a more semantically consistent multimodal representation space(Zhang et al., 2025).

In recent years, the rise of pre-trained large models has promoted the cross-modal generalization capabilities of multimodal technology. For example, OpenAI's GPT-4o can support any combination of modal input and output, and Baidu's Wenxin large model X1 Turbo has also made significant progress in modal fusion consistency and logical reasoning capabilities(Ham et al.; Xia et al., 2024). These models have strong transfer capabilities, can provide general semantic prior knowledge for user behavior analysis, and accelerate fine-tuning convergence in specific task scenarios(Kodipalli et al., 2023; Thara et al., 2022). However, most general multimodal models still face challenges in specific applications: on the one hand, they generally lack the ability to model the temporal evolution of user behavior, and it is difficult to cope with the short video environment where interests change dynamically over time; on the other hand, the modeling depth of the complex relationship between audio emotions, visual content and text semantics is limited, which can easily lead to one-sided semantic understanding(Ermishov & Savchenko, 2024).

The MT-DQN model proposed in this paper inherits the advantages of the Transformer architecture and integrates the temporal perception mechanism and the first-layer cross-modal attention network. It not only supports the semantic modeling of short video text and audio content, but also combines the user's historical behavior and social relationship characteristics to dynamically extract the law of behavior evolution. The design of this module overcomes the limitations of early fusion methods in dynamic perception and semantic coordination, and provides a more robust and realistic multimodal fusion solution for user behavior modeling in social media scenarios.

**2.2 Reinforcement Learning Modeling**

Reinforcement learning is centered on dynamic decision-making and demonstrates unique advantages in simulating the interaction between users and platforms. The classic deep Q network (DQN) combines deep learning with Q-learning, and uses the powerful function approximation ability of deep neural networks to approximate the Q-value function, thus breaking through the computational and storage bottlenecks caused by the need to store and update massive state-action pairs of Q values in high-dimensional state space(G. Martín et al., 2021; Shiau et al., 2024). For example, in some simple game scenarios, DQN uses the experience replay mechanism to randomly sample the experience of the agent's interaction with the environment for training, break data correlation, and use the target network to stabilize the target Q value, reduce training fluctuations, and enable the agent to efficiently learn the optimal decision-making strategy(Zhang et al., 2024). In the Atari game "Pong", the DQN model based on the CNN structure can accurately control the movement of the racket and achieve a higher game score(Shaheen et al., 2025).

The Proximal Policy Optimization (PPO) algorithm is an improved version of the policy gradient algorithm. By introducing a clipping function, it effectively balances the update amplitude of the policy when updating the policy, avoiding performance degradation caused by too fast policy updates(Jin & Zhang, 2024). Compared with the traditional policy gradient algorithm, it has significant improvements in training efficiency and sample utilization, and can converge to a better strategy faster with limited sample data(Ermishov & Savchenko, 2024). The hierarchical reinforcement learning model (HRL) breaks down the user decision-making process into multiple levels of sub-tasks. The high-level strategy is responsible for planning long-term goals and macro decisions, while the low-level strategy focuses on specific execution steps. It performs well in complex tasks such as robot path planning(Wang et al., 2023; H.-T. Wu et al., 2024). Existing reinforcement learning models still have many shortcomings when applied to modeling social media user decision-making behaviors. For example, the design of reward functions relies heavily on manual experience, making it difficult to accurately balance short-term benefits and long-term user value(Dulac-Arnold et al., 2021; Nasiriany et al., 2022). In addition, most models lack the ability to model the temporal dependence of user behavior and the influence of social networks, making it difficult to effectively deal with complex decision-making scenarios caused by sudden changes in user interests or social communication, and unable to fully utilize the social relationships between users and the evolution of behavior in time series to optimize decision-making strategies(Antoniadi et al., 2021).

This paper combines the reinforcement learning module with the time-series graph neural network module, optimizes the reward mechanism based on the dynamic sequence of user behavior and social relationships, and enables the model to adaptively learn user decision logic, providing a new path to solve the above problems.

## 3 Methodology

The MT-DQN (Multimodal Temporal Deep Q-Network) model takes "multimodal information perception - temporal behavior modeling - reinforcement learning decision" as its core logic and builds an end-to-end social media user decision-making behavior analysis framework. Its overall architecture is shown in Figure 1. The model comprises three interdependent and mutually reinforcing modules: a Transformer-based multimodal fusion module, a TGNN, and a DQN module. Each module serves a distinct yet complementary function, and together they form a cohesive decision-making loop through continuous data and information exchange.

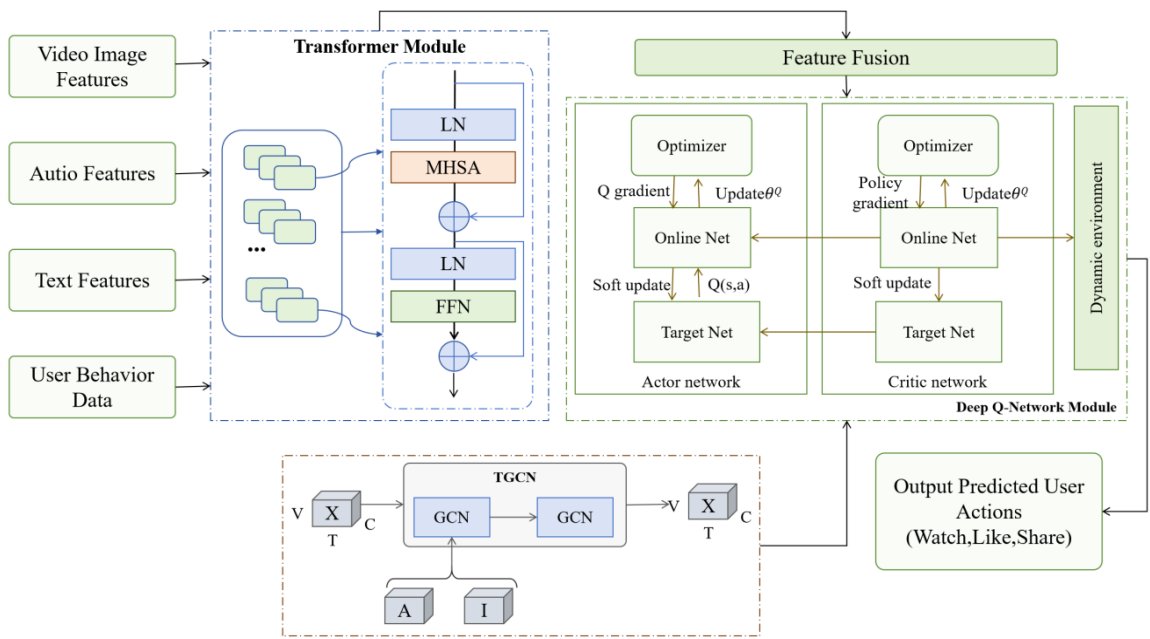

**Figure 1**.**The Overall Architecture of the MT-DQN Model for Social Media User Decision Behavior Analysis, Integrating Multimodal Fusion, Temporal Behavior Modeling, and Reinforcement Learning Decision Optimization.**

At the input layer, the model receives two core types of data from the short-video platform: (1) multimodal content data, including video images, audio streams, and text information, and (2) user behavior data, such as viewing histories, likes, comments, and other forms of interaction. The multimodal fusion module first conducts cross-modal feature extraction and integration by leveraging the self-attention mechanism inherent in the Transformer architecture, thereby enabling the model to automatically learn the importance weights associated with different modalities. These are then converted into feature vectors rich in semantic information. The temporal graph neural network module constructs a dynamic graph structure with timestamps representing users, videos, and their interaction behaviors. Using graph convolution operations and temporal attention mechanisms, this module captures the temporal dependencies in user behavior and the social network characteristics. Finally, the feature vectors output from the multimodal fusion and temporal graph neural network modules are concatenated or fused and serve as input to the deep Q-network module. In the deep Q-network module, user behaviors are modeled using both a policy network and a value network. Combined with the dynamically designed reward function, the model learns the optimal decision-making strategy using reinforcement learning algorithms, predicting the user's future behaviors, such as watching, liking, or sharing.

Through the seamless integration of these three modules, the MT-DQN model realizes a complete and adaptive pipeline—from multimodal content understanding and dynamic user behavior modeling to reinforcement learning-based decision optimization. This architecture not only overcomes the limitations of traditional models in unimodal analysis or static modeling but also continuously refines the model's ability to predict complex user decision behaviors through the feedback mechanism of reinforcement learning, providing technical support for personalized recommendation and user growth strategies on short-video platforms.

**3.1 Multimodal Fusion Module**

The multimodal fusion module serves as the central component of the MT-DQN model responsible for analyzing the semantic content of short videos, with its architectural design depicted in Figure 2. This module employs a Transformer-based encoder architecture as the core model structure, chosen for two main reasons: first, the self-attention mechanism of the Transformer overcomes the temporal dependency limitations of traditional Recurrent Neural Networks (RNNs), efficiently capturing long-range dependencies between multimodal data(Boyapati et al., 2024; Li et al., 2024); second, the multi-head attention mechanism allows for the interactive learning of modal features from multiple perspectives, enabling more precise extraction of semantic relationships between images, text, and audio compared to traditional concatenation or shallow fusion methods(Xia et al., 2022). Based on this foundation, the module follows a complete process of "data preprocessing - cross-modal feature interaction - adaptive weight fusion," transforming heterogeneous data into a unified multimodal feature vector that provides the basis for subsequent user behavior modeling.

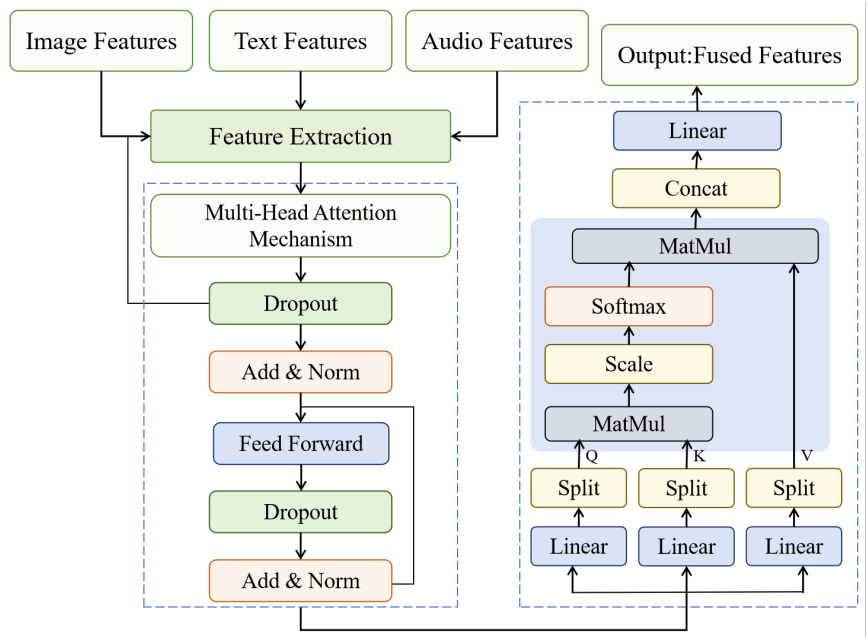

**Figure 2.** Architecture of the Multimodal Fusion Module in the MT-DQN Model for Short Video Semantic Information Processing and Feature Integration.

In the data preprocessing stage, the model receives three types of raw data from different modalities: video images $\mathbf{I} \in \mathbb{R}^{H \times W \times C}$ (height $H$, width $W$, channels $C$), text sequences $\mathbf{T} = [t_1, t_2, \ldots, t_N]$ (length $N$), and audio spectral features $\mathbf{A} \in \mathbb{R}^{F \times T}$ (frequency dimension $F$, time dimension $T$). The model employs differentiated feature extraction strategies for each modality:

- Image Feature Extraction: Visual features $\mathbf{v} \in \mathbb{R}^{D_v}$ are extracted using a pre-trained Convolutional Neural Network (ResNet-50) to capture visual information such as objects and colors in the video;

- Text Feature Extraction: The text sequence is encoded into word vectors $\mathbf{e}_t \in \mathbb{R}^{D_t}$ using a pre-trained BERT model, and average pooling is applied to obtain the text semantic feature $\mathbf{t} \in \mathbb{R}^{D_t}$;

- Audio Feature Extraction: The audio data is first transformed into a feature matrix using Mel-spectrograms, and sequential features $\mathbf{a} \in \mathbb{R}^{D_a}$ are extracted using an LSTM network.

Since the feature dimensions of each modality differ, the model uses the following linear transformations to map them to a unified dimension $D$. The calculation process is shown in Equation (1) - Equation (3):

$$\mathbf{v}' = \mathbf{W}_v \mathbf{v} + \mathbf{b}_v \tag{1}$$

$$\mathbf{t}' = \mathbf{W}_t \mathbf{t} + \mathbf{b}_t \tag{2}$$

$$\mathbf{a}' = \mathbf{W}_a \mathbf{a} + \mathbf{b}_a \tag{3}$$

where $\mathbf{W}_*$ represents the weight matrix for dimension conversion, $\mathbf{b}_*$ is the bias vector, and $\mathbf{v}', \mathbf{t}', \mathbf{a}' \in \mathbb{R}^D$ are the preprocessed visual, text, and audio feature vectors, respectively.

In the cross-modal feature interaction phase, the preprocessed multimodal features $\mathbf{v}', \mathbf{t}', \mathbf{a}'$ are input into the Transformer encoder. The core component of the Transformer, the multi-head attention mechanism (MHA), calculates the attention weights between different modality features, enabling dynamic interaction of information. For each attention head $h$, the calculation is as Equation (4):

$$\text{Attention}(\mathbf{Q}, \mathbf{K}, \mathbf{V}) = \text{softmax}\left(\frac{\mathbf{Q}\mathbf{K}^T}{\sqrt{d_k}}\right)\mathbf{V} \tag{4}$$

where $\mathbf{Q}, \mathbf{K}, \mathbf{V}$ represent the query, key, and value matrices, generated from each modality's features through linear transformations; $d_k$ is the dimension of the key vectors, used to scale the dot product results to prevent gradient vanishing issues. The multi-head attention mechanism concatenates and linearly transforms the outputs of each head, as Equation (5):

$$\text{MHA}(\mathbf{X}) = \text{Concat}(\text{head}_1, \ldots, \text{head}_H)\mathbf{W}_O \tag{5}$$

where $\text{head}_h$ is the output of the $h-th$ attention head, and $\mathbf{W}_O$ is the output weight matrix. This operation allows the model to capture relationships between modalities from multiple perspectives, enhancing the expressive power of the features.

The gating weight for the visual modality is calculated using a Sigmoid function, as Equation (6):

$$\mathbf{g}_v = \sigma(\mathbf{W}_{gv}\mathbf{v}' + \mathbf{W}_{gt}\mathbf{t}' + \mathbf{W}_{ga}\mathbf{a}' + \mathbf{b}_g) \tag{6}$$

Next, based on complementary relationships, the gating weights for the text and audio modalities $\mathbf{g}_t$ and $\mathbf{g}_a$ are derived, as Equation (7) and Equation (8):

$$\mathbf{g}_t = 1 - \mathbf{g}_v - \mathbf{g}_a \tag{7}$$
$$\mathbf{g}_a = 1 - \mathbf{g}_v - \mathbf{g}_t \tag{8}$$

The fused feature $\mathbf{f}$ is generated by element-wise weighted summation, The calculation process is shown in Equation (9):

$$\mathbf{f} = \mathbf{g}_v \odot \mathbf{v}' + \mathbf{g}_t \odot \mathbf{t}' + \mathbf{g}_a \odot \mathbf{a}' \tag{9}$$

where $\sigma$ denotes the Sigmoid function, $\mathbf{g}_*$ are the gating weight vectors for each modality, and $\odot$ represents element-wise multiplication.

Through the above design, the output multimodal feature vector $\mathbf{f}$ not only retains the unique semantics of each modality, but also integrates the complementary information across modalities, providing high-quality input representation for the subsequent temporal graph neural network module, supporting accurate modeling and decision prediction of user behavior.

**3.2 Temporal Graph Neural Network Module**

The temporal graph neural network module is the core component of the MT-DQN model that captures the dynamic changes in user behavior and social structure characteristics. Its architecture design is shown in Figure 3. This module uses TGCN as the core model structure for two key reasons: First, traditional GNNs are limited to static graph structures, making it challenging to capture the dynamic evolution of user behaviors over time. TGCN, however, extends graph convolution to aggregate features from neighboring nodes on dynamic graphs with timestamps, modeling interactions between users and videos, as well as user-user relationships and temporal dependencies(Li et al., 2023). Second, unlike methods that treat temporal information and graph structures separately, TGCN jointly models temporal and spatial dimensions, offering a more accurate representation of real-world scenarios where user decisions are shaped by both past behaviors and social network interactions(Luo et al., 2023). This architecture allows the module to learn both the social propagation patterns of user behavior and the evolving trends in user interests, providing a solid foundation for subsequent reinforcement learning decisions.

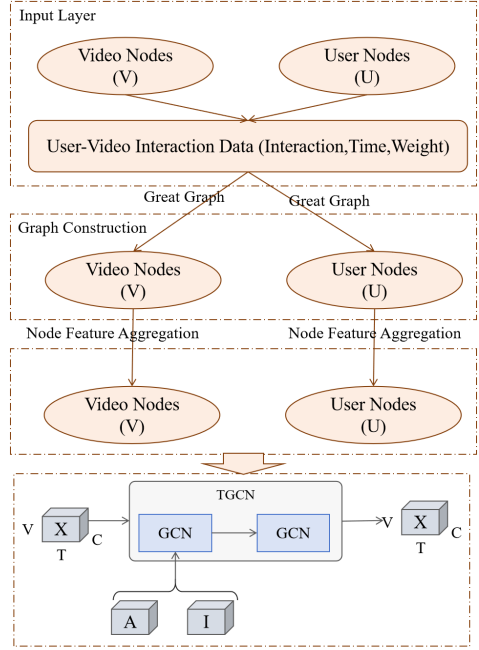

**Figure 3**. Architecture of the Temporal Graph Neural Network (TGNN) Module in the MT-DQN Model for Capturing User Behavior Dynamics and Social Structure Characteristics.

In the graph structure construction phase, the module abstracts the user behavior data into a directed, weighted graph $G=(V,E)$, where the node set $V=\{v_1,v_2,\ldots,v_{N+M}\}$ consists of $N$ user nodes and $M$ video nodes. The edge set $E$ represents the interaction relationships between the nodes (e.g., viewing, liking, commenting). Each edge $(u,v) \in E$ carries a timestamp $t$ and an interaction weight $w$ (such as viewing duration, number of likes), which characterize the temporal nature and intensity of the interaction. For example, the like behavior of user $u_i$ on video $v_k$ at time $t_j$ can be represented as the edge $(u_i, v_k)$, with a timestamp $t_j$ and a weight corresponding to the quantification of the like behavior.

To extract feature information from the graph structure, the module employs an extended graph convolution operation. Traditional GCNs struggle to handle dynamic graphs with temporal information. Therefore, this module is based on the TGCN, which improves upon traditional graph convolution by incorporating temporal information. The node feature update is computed using Equation (10):

$$\mathbf{h}_v^{l+1} = \sigma\left(\sum_{u \in N(v)} \frac{1}{c_{uv}} \mathbf{W}^l \mathbf{h}_u^l + \mathbf{b}^l\right) \tag{10}$$

where $\mathbf{h}_v^l$ represents the feature vector of node $v$ at the $l$-th layer, $N(v)$ is the set of neighbor nodes of node $v$, $c_{uv}$ is the normalization constant, $\mathbf{W}^l$ and $\mathbf{b}^l$ are the weight matrix and bias vector at layer $l$, and $\sigma$ is the activation function (e.g., ReLU). .

Temporal attention mechanism calculates attention weights for different time steps, dynamically adjusting the focus on past behaviors. The calculation of the attention weight $\alpha_t$ is given by Equation (11):

$$\alpha_t = \frac{\exp(\text{score}(\mathbf{h}_t))}{\sum_{s=1}^{T} \exp(\text{score}(\mathbf{h}_s))} \tag{11}$$

where $\text{score}(\mathbf{h}_t) = \mathbf{q}^T \tanh(\mathbf{W}\mathbf{h}_t + \mathbf{b})$ is the attention score function, and $\mathbf{q}$, $\mathbf{W}$, and $\mathbf{b}$ are learnable parameters.

The module's output temporal feature vector $\mathbf{h}_{seq}$ is obtained by a weighted summation, as is shown in Equation (12):

$$\mathbf{h}_{seq} = \sum_{t=1}^{T} \alpha_t \mathbf{h}_t \tag{12}$$

This vector integrates the temporal dependency of user behavior and social network characteristics, and serves as the input of the deep Q network module to provide a dynamic behavior pattern basis for the prediction of user decision-making behavior. Through the above design, the temporal graph neural network module effectively makes up for the shortcomings of traditional methods in modeling the temporal nature of user behavior and social relationships, and cooperates with the multimodal fusion module to provide comprehensive feature representation capabilities for the MT-DQN model.

### 3.3 Deep Q-Network Module

The Deep Q-Network (DQN) module serves as the core decision-making unit in the MT-DQN model, responsible for converting the feature representations from both the multimodal fusion and TGNN modules into optimal decision strategies for user behavior. The architecture of this module is illustrated in Figure 4. Building on the classic DQN framework, this module introduces adaptive adjustments to suit the specific characteristics of user decision-making on short-video platforms. The DQN architecture is chosen for its ability to approximate the Q-value function using deep neural networks, addressing the dimensionality challenges that traditional Q-learning encounters in high-dimensional state spaces(Yuan et al., 2022). Furthermore, incorporating target networks and experience replay mechanisms improves the stability and convergence of the training process, making DQN particularly effective for handling the dynamic nature and strong correlations in user behavior data(Xu et al., 2023).

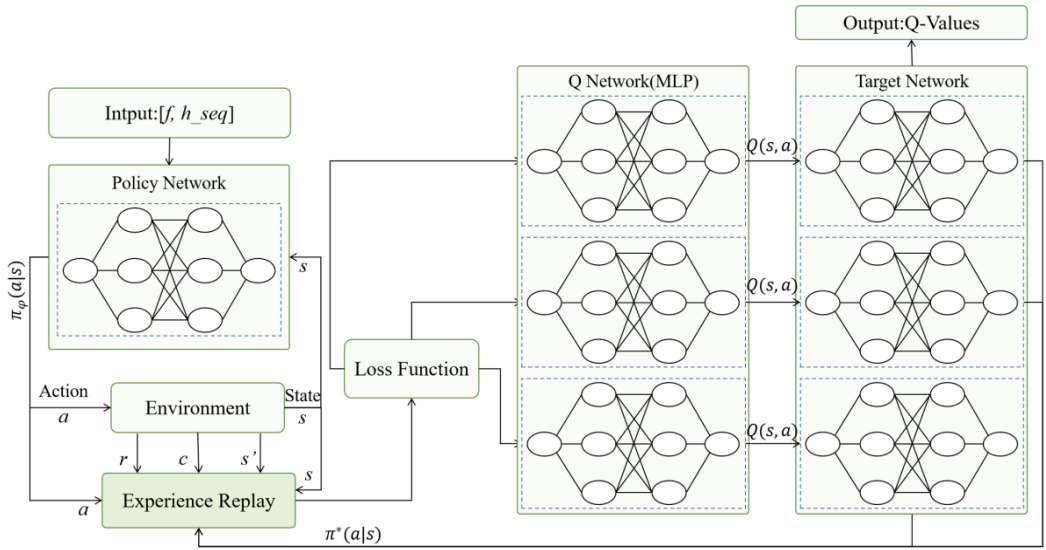

**Figure 4. Architecture of the Deep Q-Network (DQN) Module in the MT-DQN Model for Decision Making Based on Multimodal and Temporal Features.**

The module's input is the concatenation of the multimodal fusion feature $\mathbf{f}$ and the temporal feature vector $\mathbf{h}_{seq}$, resulting in the state vector $\mathbf{s} = [\mathbf{f}; \mathbf{h}_{seq}]$. This state vector fully encodes the semantic content of the short video, the temporal dependencies of user behavior, and social relationship information. The policy network $Q(\mathbf{s}, a; \theta)$ takes $\mathbf{s}$ as input and calculates the Q-values for each possible action $a \in A$ (where $A$ is the action space, such as watching, liking, sharing, etc.). The calculation process is shown in Equation (13):

$$Q(\mathbf{s}, a; \theta) = \mathbf{W}_2 \sigma(\mathbf{W}_1 \mathbf{s} + \mathbf{b}_1) + \mathbf{b}_2 \tag{13}$$

where $\mathbf{W}_1, \mathbf{W}_2$ are weight matrices, $\mathbf{b}_1, \mathbf{b}_2$ are bias vectors, and $\sigma$ is the activation function (ReLU). $\theta = \{\mathbf{W}_1, \mathbf{W}_2, \mathbf{b}_1, \mathbf{b}_2\}$ are the network parameters. The Q-value reflects the expected cumulative reward the agent will receive after executing action $a$ in state $\mathbf{s}$, with higher values indicating more favorable actions.

To optimize the policy network parameters $\theta$, the module uses the Temporal Difference (TD) learning algorithm, which iteratively updates the Q-values by minimizing the loss function $L(\theta)$. The loss function is defined in Equation (14):

$$L(\theta) = \mathsf{E}_{(\mathbf{s},a,r,\mathbf{s}') \sim D} \left[ \left( r + \gamma \max_{a'} Q(\mathbf{s}', a'; \theta^-) - Q(\mathbf{s}, a; \theta) \right)^2 \right] \tag{14}$$

where $D$ is the experience replay buffer, which stores transition samples $(\mathbf{s}, a, r, \mathbf{s}')$ (state, action, immediate reward, next state) generated from the agent's interactions with the environment; $\gamma \in [0,1]$ is the discount factor, used to balance immediate and long-term rewards; and $Q(\mathbf{s}', a'; \theta^-)$ is computed by the target network with parameters $\theta^-$, which are periodically copied from the policy network $\theta$. This separation of target network and policy network parameter update frequencies alleviates the issue of target drift during training.

In the design of the reward function $r$, the module constructs a multi-dimensional reward system that combines the platform's business objectives and user behavior characteristics. Positive rewards are given for user interactions such as likes and comments to encourage the model to recommend content the user is interested in. Negative rewards are assigned for non-interactive or early exit behaviors, prompting the model to optimize its recommendation strategy. Additionally, to prevent the model from focusing solely on short-term gains, reward components based on user retention and long-term interest stability are introduced. The specific form of the reward function is shown in Equation (15):

$$r = r_{immediate} + \lambda_1 r_{retention} + \lambda_2 r_{interest} \tag{15}$$

where $r_{immediate}$ is the immediate interaction reward, $r_{retention}$ is the user retention reward, $r_{interest}$ is the interest stability reward, and $\lambda_1, \lambda_2$ are weight coefficients used to adjust the importance of each reward component.

Through this design, the DQN module effectively maps user behavior features to decision strategies, striking a balance between optimizing immediate user responses and exploring long-term value potentials. By continuously optimizing the recommendation strategy, the module seeks to approach the optimal decision. The action probability distribution output by the module provides a directly applicable decision basis for personalized recommendations, content optimization, and other applications on short-video platforms. In collaboration with other modules, it forms a complete framework for user decision behavior analysis and prediction.

### 3.4 Inter-Module Synergy Mechanism

The three core modules of the MT-DQN model operate as an integrated system through continuous data transfer and information feedback, thereby achieving a closed-loop pipeline spanning content understanding, behavior modeling, and decision optimization. The multimodal fusion module transforms the image, text, and audio components of short videos into rich semantic feature representations, while the TGNN module extracts dynamic, temporally aware features from user behavior data. After fusion, these two types of features serve as inputs to the deep Q-network module, supporting decision-making analysis based on the dual dimensions of "content" and "behavior." For example, when a user browses a food-related short video, the multimodal fusion module provides information such as visuals and captions, while the temporal graph neural network module incorporates the user's historical browsing records and social relationships to assist the deep Q-network module in determining whether the user is likely to perform actions such as liking or sharing.

During the training process, the deep Q network module calculates the loss based on the decision results and reward signals, optimizes its own parameters, and feeds back key information to the multimodal fusion module and the time-series graph neural network module. The multimodal fusion module adjusts the cross-modal attention weights accordingly to strengthen the extraction of important modal information; the time-series graph neural network module optimizes the graph convolution and attention mechanism to capture more critical behavior patterns. This two-way collaborative mechanism enables the model to quickly adapt to data changes while continuously accumulating decision-making experience to achieve accurate prediction of user behavior.

# 4 Experiment

## 4.1 Experimental Data

This experiment uses three publicly available datasets: YouTube-8M, Allo-AVA, and Weibo User Interaction. These datasets offer a comprehensive mix of short-video content and user behavior data, providing a strong basis for model training and performance evaluation. The YouTube-8M dataset, which serves as the primary source of video data, contains 8 million short video samples across various domains such as film, music, and education. Each sample includes video frame images, audio waveforms, and text labels, with its large-scale, heterogeneous structure providing a rigorous test for the model's multimodal fusion capabilities(Abu-El-Haija et al., 2016). The Allo-AVA dataset, with over 100,000 video samples featuring audio-visual synchronization annotations, emphasizes the alignment between audio and visual elements. This dataset improves the model's ability to understand the interplay between audio and visual semantics, reducing the bias that might arise from relying on a single modality(Punjwani & Heck, 2024). The Weibo User Interaction dataset records user interactions, such as likes, comments, and reposts, on the Weibo platform, as well as the social relationship networks between users(Hu et al., 2020). In this dataset, users and videos are represented as nodes, while the edge weights and timestamps denote the intensity of interactions and the sequence of behavior occurrences.

In the data preprocessing stage, we performed a series of time alignment processes on the video data of YouTube-8M and Allo-AVA. First, in the alignment of video and audio, we adopted a precise synchronization strategy based on timestamps to ensure that each frame of video and the corresponding audio clip can be aligned in the time dimension. At the same time, in the user behavior data, we matched the timestamp with the video playback time to associate the user's likes, comments and other behaviors with the corresponding video content to ensure the temporal consistency and synchronization of multimodal data. In addition, the audio feature extraction uses Mel spectrum conversion combined with LSTM to capture the temporal features in the audio signal and synchronize it with the video frame image and text information. All data have been strictly time-aligned to ensure that the model can fully learn the correlation between different modalities during the training process. Finally, all data are divided into training set, validation set and test set in a ratio of 7:1:2. The training set is used for model parameter optimization, the validation set assists in hyperparameter adjustment, and the test set independently evaluates the model generalization ability to ensure the reliability and effectiveness of the experimental results.

## 4.2 Experimental Setup

The experiments are conducted on an NVIDIA RTX 3090 GPU to accelerate computation, supported by a hardware configuration comprising 64GB of RAM and an Intel Core i9-12900K CPU. The deep learning framework used is PyTorch 1.12, which ensures efficient handling of large-scale data processing and complex model training workflows.

Regarding model parameterization, the Transformer encoder within the multimodal fusion module is configured with 6 layers, each comprising 12 attention heads. The extracted image, text, and audio features are uniformly mapped into a 768-dimensional semantic space to facilitate integrated multimodal representation. The temporal graph convolutional layers in the temporal graph neural network module are set to 3 layers, with the number of convolutional kernels being 64, 128, and 256, respectively, to progressively extract higher-order graph structure features. The policy network of the deep Q-network module consists of 3 fully connected layers, with the number of neurons being 512, 256, and 128, respectively. The output dimension is consistent with the size of the action space.

During the training process, Adam was selected as the optimizer, the initial learning rate was set to 0.001, and the cosine annealing strategy was used to dynamically adjust the learning rate; the batch size was set to 64, and the number of training rounds was 50 rounds to balance the training efficiency and model convergence effect. To alleviate the overfitting problem, Dropout was added after the Transformer layer and the fully connected layer, and the inactivation rate was set to 0.2; the experience replay buffer capacity of the deep Q network module was set to 100,000, and a random uniform sampling strategy was used to break the time correlation between data, effectively improving sample diversity; the target network synchronized parameters with the policy network every 1,000 steps to reduce the instability caused by target drift during policy convergence; in addition, in order to further enhance the stability and generalization ability of training, a fixed Q target value mechanism and a gradient clipping strategy were introduced during the training process to prevent value function estimation oscillation and gradient explosion, respectively. The discount factor $\gamma$ in reinforcement learning was set to 0.95 to reasonably balance immediate rewards and long-term benefits, ensuring that the model can achieve stable and efficient training and optimization in user behavior prediction tasks.

**4.3 Evaluation Metrics**

To comprehensively evaluate the performance of the MT-DQN model in user behavior prediction and decision optimization tasks, this experiment employs evaluation metrics drawn from four key dimensions: prediction accuracy, ranking quality, error magnitude, and model stability.

The F1 score is used to assess the model's precision and recall capabilities in predicting user behavior. The F1 score balances the model's ability to recognize both positive and negative samples by combining precision and recall. Its calculation formula is as Equation (16) :

$$F1 = 2 \times \frac{\text{Precision} \times \text{Recall}}{\text{Precision} + \text{Recall}} \tag{16}$$

where precision $\text{Precision} = \frac{TP}{TP+FP}$ and recall $\text{Recall} = \frac{TP}{TP+FN}$. $TP$ represents true positives, $FP$ represents false positives, and $FN$ represents false negatives.

Normalized Discounted Cumulative Gain (NDCG) is introduced to evaluate how well the model-generated user behavior recommendation list matches the true preferences. The calculation process is shown in Equation (17) :

$$\text{NDCG}_k = \frac{\text{DCG}_k}{\text{IDCG}_k} \tag{17}$$

where $k$ is the length of the recommendation list, $\text{DCG}_k = \sum_{i=1}^{k} \frac{2^{r_i} - 1}{\log_2(i+1)}$, and $r_i$ is the relevance score of the $i$-th recommendation. $\text{IDCG}_k$ is the ideal DCG value in an optimal situation.

Mean Squared Error (MSE) is used to quantify the deviation between the predicted Q-values and the actual cumulative rewards. The calculation process is shown in Equation (18) :

$$\text{MSE} = \frac{1}{n}\sum_{i=1}^{n}(y_i - \hat{y}_i)^2 \tag{18}$$

where $n$ is the number of samples, $y_i$ is the actual cumulative reward, and $\hat{y}_i$ is the predicted Q-value by the model. A smaller MSE indicates more accurate predictions of the decision value in user behavior.

Mean Absolute Error (MAE) is used to evaluate the fluctuation of prediction results across different test datasets. The calculation process is shown in Equation (19) :

$$\text{MAE} = \frac{1}{n}\sum_{i=1}^{n}|y_i - \hat{y}_i| \tag{19}$$

MAE provides an intuitive measure of the average deviation between predicted and actual values.

In addition, in order to more intuitively measure the practicality of the model recommendation ranking, this study also introduces the hit rate as a supplementary indicator to evaluate the proportion of real user behavior samples successfully hit by the model in the recommendation list. The hit rate is particularly suitable for Top-K recommendation scenarios and can reflect the model's ability to rank the content that users are really interested in(Bajusz & Keserű, 2022). Although the hit rate has not yet been used as a unified benchmark indicator in some recommendation system studies, it has a high reference value in the industry and Top-K performance evaluation practice.

Through the above five-dimensional indicator settings, this paper systematically evaluates the comprehensive performance of the MT-DQN model in processing user behavior data, providing a scientific and comprehensive quantitative basis for model optimization and comparative experiments.

**4.4 Ablation Experiment**

To verify the effectiveness of the core components of Transformer, TGNN, and DQN in the MT-DQN model, this experiment builds a simplified model by gradually removing each module or replacing key mechanisms, and compares its performance on three datasets: YouTube-8M, Allo-AVA, and Weibo. The experiment keeps other parameters the same and only adjusts the target components to quantify the contribution of each module to the overall model performance. The experimental results are shown in Table 1.

Table 1 Comparison of F1 Score, NDCG@5, MSE, and MAE Performance in the Ablation Study of MT-DQN Core Components (Transformer, TGNN, DQN) on the YouTube-8M, Allo-AVA, and Weibo User Interaction Datasets.

| Model Variant | YouTube-8M Dataset | | | | Allo-AVA Dataset | | | | Weibo User Interaction Dataset | | | |
|---|---|---|---|---|---|---|---|---|---|---|---|---|
| | F1 | NDCG@5 | MSE | MSE | F1 | NDCG@5 | MSE | MSE | F1 | NDCG@5 | MSE | MSE |
| **MT-DQN** | **0.832** | **0.885** | **0.089** | **0.227** | **0.805** | **0.862** | **0.098** | **0.239** | **0.853** | **0.898** | **0.082** | **0.215** |
| -Transformer | 0.751 | 0.801 | 0.132 | 0.295 | 0.723 | 0.781 | 0.153 | 0.746 | 0.782 | 0.831 | 0.128 | 0.279 |
| -TGNN | 0.778 | 0.823 | 0.121 | 0.281 | 0.746 | 0.804 | 0.142 | 0.303 | 0.801 | 0.849 | 0.117 | 0.265 |
| -DQN | 0.743 | 0.792 | 0.141 | 0.302 | 0.718 | 0.776 | 0.158 | 0.321 | 0.771 | 0.824 | 0.135 | 0.288 |

The results in Table 1 reveal that removing the Transformer module has the most significant negative impact on model performance, with a 7.2% decrease in F1 score and a 6.8% drop in NDCG@5 across all three datasets. This highlights the Transformer's crucial role in multimodal feature fusion and semantic understanding. Without it, the model struggles to capture the intricate relationships between images, text, and audio through self-attention, leading to a reduction in feature expressiveness and negatively affecting behavior prediction and decision-making.

Removing the TGNN module results in a 5.4% drop in F1 score and a 5.2% decrease in NDCG@5, underscoring its importance in modeling the temporal dynamics of user behavior and social network structures. For example, in the Weibo dataset, where user interactions are frequent and exhibit strong temporal dependencies, omitting TGNN hampers the model's ability to capture these patterns and social contagion effects, leading to a noticeable decline in accuracy for predicting future user behavior.

Excluding the DQN module causes a significant increase in MSE and MAE, along with a 7.1% reduction in F1 score, confirming its essential role in optimizing decision-making within the MT-DQN model. Without DQN, the model cannot effectively learn optimal decision strategies through reinforcement learning, making it difficult to balance short-term rewards with long-term user behavior outcomes. This results in a marked decline in both decision-making accuracy and stability.

The Transformer, TGNN, and DQN modules in the MT-DQN model are interdependent and indispensable. The Transformer is responsible for deep multimodal fusion and semantic extraction, TGNN models the temporal and structural aspects of user behavior, and DQN refines decision strategies based on environmental feedback. Together, these components enable the model to excel in predicting complex user behaviors and optimizing decision-making.

**4.5 Result and Analysis**

To evaluate the MT-DQN model's effectiveness, this study compares it with six models, including traditional methods, established frameworks, and recent innovations, forming a comprehensive benchmark system. Among these, Concat-Modal (a traditional concatenation model) serves as the baseline for multimodal fusion, demonstrating the need for deep fusion mechanisms. Single-LSTM uses only text to model sequential behavior, emphasizing the value of multimodal data. Vanilla-DQN (classic deep Q-network) lacks multimodal and temporal graph structures, highlighting the benefits of the extended modules. GraphSAGE+GRU combines graph sampling with recurrent networks, representing traditional graph-sequence modeling. MHA-Fusion uses multi-head attention to fuse multimodal features but omits temporal graph networks, isolating the effects of cross-modal interaction and temporal modeling. Temporal-DQN adds simple temporal features to DQN, showcasing the limitations of basic temporal models. Through these comparative models, we can systematically analyze the advantages of MT-DQN in multimodal fusion depth, temporal modeling accuracy, and decision optimization ability. The experimental results are shown in Table 2.

Table 2 Performance Comparison of the MT-DQN Model with Traditional Multimodal Models, Reinforcement Learning Models, and Modified Variants on YouTube-8M, Allo-AVA, and Weibo Datasets, based on F1 Score, NDCG@5, MSE, and MAE.

| Model | YouTube-8M Dataset | | | | Allo-AVA Dataset | | | | Weibo User Interaction Dataset | | | |
|---|---|---|---|---|---|---|---|---|---|---|---|---|
| | F1 | NDCG@5 | MSE | MSE | F1 | NDCG@5 | MSE | MSE | F1 | NDCG@5 | MSE | MSE |
| **MT-DQN** | **0.832** | **0.885** | **0.089** | **0.227** | **0.805** | **0.862** | **0.098** | **0.239** | **0.853** | **0.898** | **0.082** | **0.215** |
| Concat-Modal(Nuthakki et al., 2023) | 0.721 | 0.763 | 0.153 | 0.312 | 0.703 | 0.741 | 0.162 | 0.321 | 0.737 | 0.774 | 0.149 | 0.305 |
| Single-LSTM(Zhi, 2024) | 0.689 | 0.732 | 0.171 | 0.335 | 0.668 | 0.715 | 0.183 | 0.347 | 0.712 | 0.753 | 0.168 | 0.326 |
| Vanilla-DQN(J. Wu et al., 2024) | 0.745 | 0.781 | 0.138 | 0.298 | 0.732 | 0.776 | 0.145 | 0.306 | 0.758 | 0.792 | 0.132 | 0.289 |
| GraphSAGE+GRU(Yuan et al., 2025) | 0.763 | 0.795 | 0.129 | 0.287 | 0.748 | 0.789 | 0.134 | 0.293 | 0.769 | 0.801 | 0.127 | 0.283 |
| MHA-Fusion(Liu et al., 2025) | 0.772 | 0.803 | 0.125 | 0.282 | 0.756 | 0.797 | 0.131 | 0.288 | 0.777 | 0.809 | 0.123 | 0.278 |
| Temporal-DQN(Xu et al., 2019) | 0.758 | 0.791 | 0.136 | 0.294 | 0.741 | 0.784 | 0.141 | 0.301 | 0.764 | 0.798 | 0.130 | 0.286 |

According to the results presented in Table 2, the MT-DQN model significantly outperforms all comparative models across key performance metrics on the three evaluated datasets, demonstrating the comprehensive advantages of its technical framework. Specifically, regarding multimodal fusion capability, the F1 score of MT-DQN is, on average, 10.97% higher than that of Concat-Modal and 5.83% higher than that of MHA-Fusion, indicating that the Transformer's self-attention mechanism can more effectively capture deep semantic relationships among images, text, and audio, compared to simple concatenation or shallow attention mechanisms. For instance, on the YouTube-8M dataset, the F1 score of MT-DQN is 0.832, while that of Concat-Modal is only 0.721. This gap primarily arises from the inability of traditional concatenation to model the dynamic dependencies between modalities, leading to redundant feature representation and lack of semantic hierarchy.

In terms of temporal behavior modeling, MT-DQN's NDCG@5 is on average 8.12% higher than that of GraphSAGE+GRU and 9.36% higher than that of Temporal-DQN. This demonstrates that the temporal graph convolution mechanism of TGNN, compared to traditional LSTM or simple temporal feature stacking, can more precisely capture the time-series patterns of user behavior and the social network propagation models. Taking the Weibo dataset as an example, the strong temporal nature of user interactions requires the model to possess long-range dependency modeling capabilities. MT-DQN's NDCG@5 reaches 0.898, significantly outperforming GraphSAGE+GRU's 0.801, confirming the irreplaceability of TGNN in dynamic graph structure modeling.

In terms of decision optimization, MT-DQN's MSE and MAE are reduced by an average of 34.8% and 26.5% compared to Vanilla-DQN. This shows that the combination of deep Q networks with multimodal fusion and temporal graph modeling effectively alleviates the generalization problem in high-dimensional state spaces in traditional reinforcement learning. For example, on the Allo-AVA dataset, MT-DQN's MSE is 0.098, while Vanilla-DQN's is 0.145. The difference arises because MT-DQN uses joint inputs of cross-modal features and behavioral temporal features, enabling the Q-value function to more accurately approximate the long-term value of user behavior.

It is worth noting that in the Allo-AVA dataset, although the overall prediction error is low, there are abnormal fluctuations in the model prediction performance for some specific types of videos (such as samples with high-noise audio tracks or asynchronous audio and video content), and the MSE of individual categories is significantly higher than the average level of the dataset. Further analysis found that such samples have modal information conflicts or situational understanding biases in multimodal representation, which causes the model to generate misleading attention distribution during fusion, thereby affecting the Q-value prediction accuracy of the policy network. In addition, the audio emotion labels of some videos in Allo-AVA are relatively subjective and lack clear semantic anchors, which also increases the difficulty of cross-modal alignment. In the future, we will consider introducing a modal confidence mechanism or hedging mechanism to dynamically adjust the influence of each modality on the policy output to further improve the model's robustness to noisy samples and its adaptability to abnormal scenarios.

The performance advantage of MT-DQN arises from the synergistic effect of its three modules: Transformer lays the foundation for multimodal semantic understanding, TGNN enhances behavioral dynamic modeling, and DQN optimizes decision strategies based on environmental feedback. This combination not only overcomes the limitations of unimodal or traditional modeling methods but also provides an optimal solution for user behavior prediction and decision-making tasks through deep fusion and dynamic modeling.

In Figure 5, we illustrate the trade-offs between recommendation relevance (measured by metrics such as NDCG/F1 score) and recommendation diversity (measured by internal list similarity) for MT-DQN compared to models like Concat-Modal, Vanilla-DQN, and MHA-Fusion. The x-axis represents the recommendation relevance of each model, while the y-axis represents recommendation diversity, with each point corresponding to a different model.

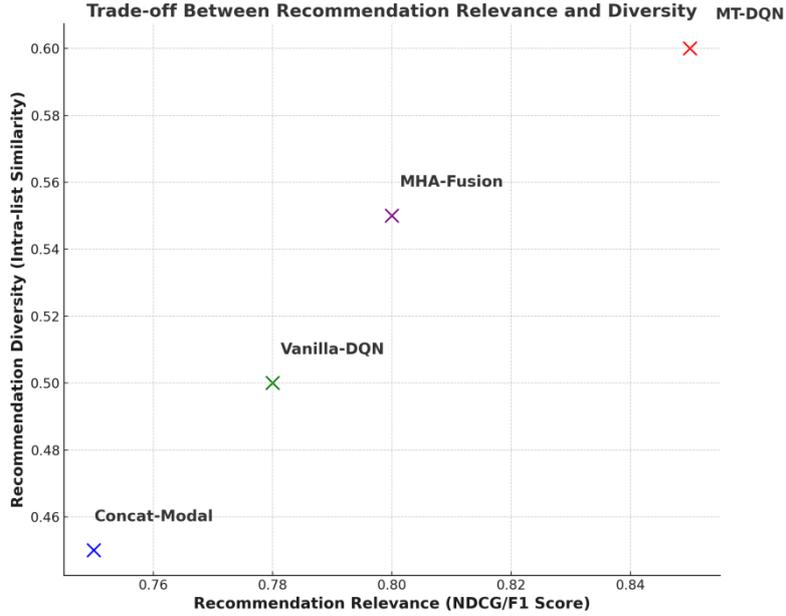

**Figure 5**. **The trade-off between recommendation relevance and diversity: Visualization of MT-DQN and the comparison model.**

As illustrated in Figure 5, MT-DQN excels in both recommendation relevance and diversity. Compared to competing models, MT-DQN achieves high recommendation relevance while simultaneously maintaining a high degree of diversity. This dual strength is particularly critical, as recommendation systems must balance aligning content with user interests while avoiding content monotony and offering a broader range of engaging options. The Concat-Modal model, while having some advantages in diversity, has lower relevance, indicating that it processes multi-modal data through simple concatenation, which makes it difficult to deeply capture the relationships between different modalities, leading to poorer recommendation accuracy. Vanilla-DQN and MHA-Fusion have improved relevance but perform less well in diversity compared to MT-DQN, suggesting that these models, while optimizing relevance, fail to effectively balance diversity. By integrating the deep fusion mechanism of Transformer and TGNN, MT-DQN can optimize both the accuracy and diversity of recommendations, ensuring that the recommendation list not only aligns with user interests but also avoids content repetition, enhancing user experience. This advantage highlights the potential of MT-DQN in recommendation systems for short video platforms, especially in scenarios where large amounts of heterogeneous data need to be processed and personalized, diverse recommendations are required.

Figure 6 presents the hit rate comparison between MT-DQN and other models, including Concat-Modal, Vanilla-DQN, and MHA-Fusion, across different recommendation ranking positions (Top-1, Top-3, Top-5). The x-axis denotes the evaluated models, while the y-axis indicates the recommendation hit rate, defined as the proportion of true positive samples correctly identified within the recommendation list.

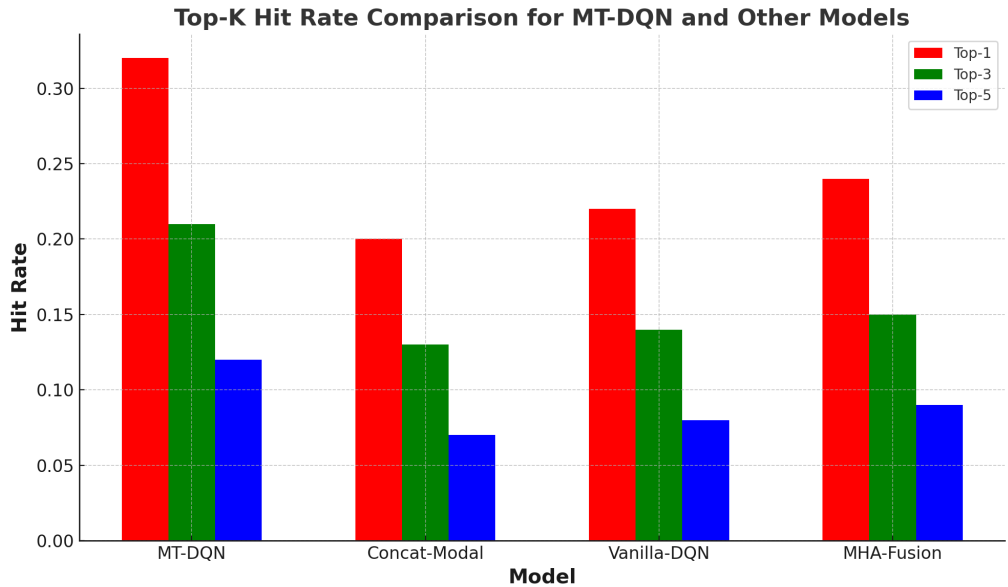

Figure 6. **Performance of MT-DQN and the comparison model in Top-K recommendation hit rate.**

As depicted in Figure 6, MT-DQN demonstrates superior performance across all ranking positions, particularly at the Top-1 position, where it achieves a hit rate of 32%, markedly outperforming other models. This result underscores MT-DQN's exceptional accuracy in recommendation sorting. Notably, compared to Concat-Modal, MT-DQN's Top-1 hit rate improves by 60%, highlighting its substantial advantage in prioritizing highly relevant content aligned with user interests. Although Vanilla-DQN and MHA-Fusion have seen improvements in hit rates, they still lag behind MT-DQN in the Top-1 position, especially in the Top-3 and Top-5 rankings, where MT-DQN's performance is even more pronounced. This suggests that MT-DQN can effectively optimize the ranking of recommendation lists by deeply integrating multi-modal features and temporal behavior information, ensuring that the most relevant items are prioritized for display. In contrast, other models like Concat-Modal, which only use a simple concatenation method to integrate multi-modal data, fail to capture the deep relationships between modalities, leading to lower hit rates, particularly in the Top-1 ranking.

Figure 7 illustrates the cross-modal attention heatmap generated by the MT-DQN model, showcasing the distribution of attention weights between image feature regions and associated text/audio features. The depth of the color indicates the strength of the attention weight; the darker the color, the stronger the association between the region and the feature.

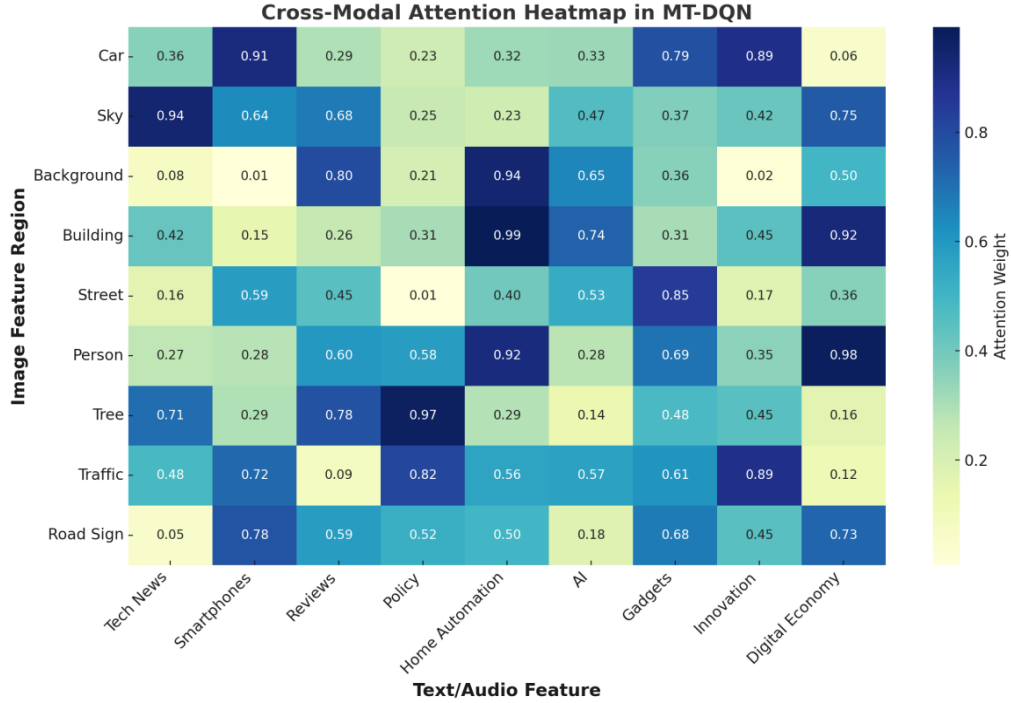

**Figure 7**. **Cross-modal attention heat map in MT-DQN model: correlation between image features and text/audio featurese.**

The results in Figure 7 show that MT-DQN performs well in capturing cross-modal relationships between image, text, and audio features, especially in semantic alignment of image and text descriptions. For example, in the "Tech News" and "Smartphone" text features, the model shows high attention weights to regions such as "car", "tree", and "street" in the image, indicating that MT-DQN can accurately identify and align image content related to text descriptions. Specifically, the "tree" region has received high attention weights in multiple text features (such as "Tech News" and "Innovation"), showing the model's high attention to this object, which indicates that the model can effectively capture text-related image semantic features when multimodal data is fused.

In addition, we further analyzed and found that "car" in the image also shows high correlation in text descriptions related to "smartphone" and "policy", indicating that the model can not only identify clear objects in the image, but also capture deep semantic relationships related to different contexts. The high correlation between "car" in the image and "smartphone" text may reflect that users tend to pay attention to other technologies or devices related to "smartphone" when searching for "technology news" content, which reveals the model's ability to capture potential user interests.

This experiment randomly selected 100 test users from the Weibo User Interaction dataset. The Top-5 recommendation lists generated by MT-DQN were compared against those produced by three representative models: Concat-Modal, Vanilla-DQN, and MHA-Fusion. For each recommended item, actual user interaction behaviors (such as likes, comments, and shares) were annotated, and the hit rate, defined as the proportion of true positive samples within the recommendation list, was calculated. The specific results are shown in Table 3.

**Table 3 Average Hit Rate and Hit Proportion at Each Position in the Top-5 Recommendation List for Different Models on the Weibo Dataset.**

| Model | Average Hit Rate | Recommendation Item 1 Hit Rate | Recommendation Item 2 Hit Rate | Recommendation Item 3 Hit Rate | Recommendation Item 4 Hit Rate | Recommendation Item 5 Hit Rate |
|---|---|---|---|---|---|---|
| **MT-DQN** | 0.68 | 32% | 21% | 12% | 8% | 5% |
| Concat-Modal | 0.45 | 20% | 13% | 7% | 3% | 2% |
| Vanilla-DQN | 0.49 | 22% | 14% | 8% | 3% | 2% |
| MHA-Fusion | 0.52 | 24% | 15% | 9% | 3% | 1% |

As summarized in Table 3, MT-DQN significantly outperforms the comparative models in both recommendation ranking accuracy and rationality. The average hit rate achieved by MT-DQN is 0.68, representing a 51% improvement over Concat-Modal. This notable gain underscores that through the synergistic integration of the Transformer and TGNN, MT-DQN more effectively captures the alignment between user interests and content relevance. In terms of recommendation position distribution, the hit rate for Recommendation Item 1 in MT-DQN is 32%, almost 1.6 times that of Concat-Modal (20%), demonstrating its ability to prioritize highly relevant content. This result corresponds with the excellent performance in NDCG@5.

**Table 4 Top-5 Recommendation List and Hit Status Comparison for User ID U001 between MT-DQN and the Comparative Models.**

| Model | Recommendation Item 1 | Recommendation Item 2 | Recommendation Item 3 | Recommendation Item 4 | Recommendation Item 5 | Hit Status |
|---|---|---|---|---|---|---|
| **MT-DQN** | In-depth Technology News ★ | New Smartphone Review ★ | Industry Tech Summit Live | Technology Policy Analysis | Smart Home Product Launch | Hit 2 items |
| Concat-Modal | Celebrity News Updates | Food Review Video | In-depth Technology News ★ | Fashion Tutorial | Car Test Drive Experience | Hit 1 items |
| Vanilla-DQN | Financial Market Analysis | Travel Guide | Technology Policy Analysis | New Smartphone Review ★ | Fitness Tutorial | Hit 1 items |
| MHA-Fusion | In-depth Technology News ★ | Beauty Product Recommendations | Digital Product Repair Guide | Sports Event Recap | Technology Conference News | Hit 1 items |

Additionally, user-specific case analysis, such as for user ID U001 (Table 4), reveals that MT-DQN places two "technology news" items in the top two recommendation positions, both of which precisely match the user's actual interaction behaviors. In contrast, the comparative models exhibit clear deficiencies. For instance, Concat-Modal ranks "celebrity news" first, while Vanilla-DQN mixes in unrelated content such as "financial market analysis" and "travel guides," reflecting the misjudgment of user interests due to the lack of multimodal deep fusion and temporal behavior modeling. This further validates that MT-DQN, by leveraging the temporal features of users' historical behaviors and the multimodal semantics of the content, can generate more relevant recommendation lists tailored to users' needs in real-world recommendation scenarios.

## 5 Conclusion and Discussion

This paper proposes the MT-DQN model, which integrates Transformer, TGNN, and DQN, components to address the complex challenge of user behavior prediction and recommendation decision-making within the short-video platforms. By combining deep multimodal features fusion (encompassing image, text, and audio data), temporal dynamics modeling of user behavior (via TGNN), and reinforcement learning-based decision optimization (via DQN), the MT-DQN model demonstrates substantial performance advantages across three public datasets (YouTube-8M, Allo-AVA, Weibo User Interaction). Experimental results show that MT-DQN improves core metrics such as F1 score and NDCG@5 by over 10% compared to traditional multimodal models (e.g., Concat-Modal) and by more than 8% compared to classic reinforcement learning models (e.g., Vanilla-DQN). Ablation studies further confirm the indispensable roles of Transformer's cross-modal attention mechanism, TGNN's temporal modeling ability, and DQN's decision optimization module in enhancing model performance. Visual analysis (such as cross-modal attention heatmaps and recommendation list comparisons) demonstrates that MT-DQN effectively aligns multimodal semantics, captures temporal patterns in user behavior, and balances recommendation relevance and diversity, providing a more accurate and interpretable solution for short-video recommendation systems.

Despite its promising experimental performance, the MT-DQN model exhibits several noteworthy limitations. First, its architectural complexity leads to substantial computational demands during training, which may present challenges for deployment in real-time recommendation scenarios where low-latency inference is critical. Second, the current study primarily focuses on individual user behavior modeling and has not fully addressed the impact of group interactions within social networks, such as group preference diffusion and social influence propagation. Additionally, the handling of multimodal data noise still requires optimization. For instance, in low-quality short-video scenarios (e.g., blurry images or noisy audio), the model's ability to capture key semantics may decrease. Lastly, although interpretability has been partially achieved through attention heatmaps, there is a lack of dynamic tracking of long-term user interest evolution, making it difficult to fully reveal the deep logic behind the model's decision-making.

In view of the above shortcomings, future research will be expanded and improved from several aspects. First, we will explore lightweight model architectures, reduce the computational complexity of MT-DQN and improve its online reasoning efficiency by introducing technologies such as knowledge distillation and model pruning to meet the needs of real-time recommendation scenarios; second, we will expand the model's ability in modeling social group behavior, and introduce graph attention networks (GAT) or social diffusion models to incorporate user social relationships and group interaction patterns into the temporal graph modeling framework to further enhance the social synergy and adaptability of the recommendation system; third, in response to the noise problem in multimodal data, combined with technologies such as adversarial training or self-supervised learning, improve the model's robustness and feature extraction accuracy in low-quality data; finally, deepen the model's interpretability research, develop a dynamic interest tracking module, and through the temporal visualization of user interest vectors, reveal the model's mechanism in capturing users' long-term preferences and short-term interest fluctuations, thereby enhancing the transparency of the recommendation system and user trust.

## Data availability statement

The data and materials used in this study are not currently available for public access. Interested parties may request access to the data by contacting the corresponding author.

## Conflict of Interests Statement

The authors declare that the research was conducted in the absence of any commercial or financial relationships that could be construed as a potential conflict of interest.

## Funding

No.

## Consent for publication

All authors of this manuscript have provided their consent for the publication of this research.